# Embedded Bayesian Network Classifiers

David Heckerman
heckerma@microsoft.com

Christopher Meek
meek@microsoft.com

**Abstract**

Low-dimensional probability models for local distribution functions in a Bayesian network include decision trees, decision graphs, and causal independence models. We describe a new probability model for discrete Bayesian networks, which we call an *embedded Bayesian network classifier* or EBNC. The model for a node $Y$ given parents $\mathbf{X}$ is obtained from a (usually different) Bayesian network for $Y$ and $\mathbf{X}$ in which $\mathbf{X}$ need not be the parents of $Y$. We show that an EBNC is a special case of a softmax polynomial regression model. Also, we show how to identify a non-redundant set of parameters for an EBNC, and describe an asymptotic approximation for learning the structure of Bayesian networks that contain EBNCs. Unlike the decision tree, decision graph, and causal independence models, we are unaware of a semantic justification for the use of these models. Experiments are needed to determine whether the models presented in this paper are useful in practice.

# 1 Introduction

Several researchers have demonstrated that Bayesian networks provide better predictions when local distribution functions (also known as conditional probability tables for domains in which all variables have a finite number of states) are modeled with a small number of parameters. Such parsimonious models include decision trees, decision graphs, and causal-independence models (e.g., Friedman and Goldszmidt, 1996; Chickering et al., 1997; Meek and Heckerman, 1997). In this paper, we introduce another parsimonious model for Bayesian networks in which each variable has a finite number of states, known as an *embedded Bayesian network classifier* or EBNC. The model for a node $Y$ given parents $\mathbf{X}$ is obtained from a (usually different) Bayesian network for $Y$ and $\mathbf{X}$ in which $\mathbf{X}$ need not be the parents of $Y$.

In Section 2 we introduce the model. In Section 3, we describe a simple Bayesian-network inference algorithm that can be used to compute the probability distribution for $Y$ given $\mathbf{X}$ as determined by an EBNC. In Section 4, we show how to identify a non-redundant set of parameters for an EBNC and consequently its dimension. In Section 5, we describe an approximation method for learning the structure of a Bayesian network that contains EBNCs. In particular, we show that a Laplace approximation can be used to approximate the marginal likelihood of a Bayesian network that contains EBNCs. The method can be used to select both models and input features for classification. In Section 6, we present a more efficient procedure for finding a set of non-redundant parameters for an EBNC. As a result, we show that an EBNC is a special case of a softmax polynomial regression model.

One word of caution is warranted. Unlike the decision tree, decision graph, and causal independence models, we are unaware of a semantic justification for the use of these models. In fact, there are theoretical reasons that suggest the use of EBNCs may be unreasonable (Heckerman and Meek, 1997). Experiments are needed to determine whether the models presented in this paper are useful in practice.

The terminology and notation we need is as follows. We denote a variable by an upper-case letter (e.g., $X, Y, X_i, \Theta$), and the state or value of a corresponding variable by that same letter in lower case (e.g., $x, y, x_i, \theta$). We denote a set of variables by a bold-face upper-case letter (e.g., $\mathbf{X}, \mathbf{Y}, \mathbf{X}_i$). We use a corresponding bold-face lower-case letter (e.g., $\mathbf{x}, \mathbf{y}, \mathbf{x}_i$) to denote an assignment of state or value to each variable in a given set. We say that variable set $\mathbf{X}$ is in *configuration* $\mathbf{x}$. We use $p(\mathbf{X} = \mathbf{x}|\mathbf{Y} = \mathbf{y})$ (or $p(\mathbf{x}|\mathbf{y})$ as a shorthand) to denote the probability or probability density that $\mathbf{X} = \mathbf{x}$ given $\mathbf{Y} = \mathbf{y}$. We also use $p(\mathbf{x}|\mathbf{y})$ to denote the probability distribution (both mass functions and density functions) for $\mathbf{X}$ given $\mathbf{Y} = \mathbf{y}$. Whether $p(\mathbf{x}|\mathbf{y})$ refers to a probability, a probability density, or a probability distribution will be clear from context.

We use $\mathbf{m}$ and $\boldsymbol{\theta}_m$ to denote the structure and parameters of a model, respectively. When $(\mathbf{m}, \boldsymbol{\theta}_m)$ is a Bayesian network for variables $\mathbf{Z}$, we write the usual factorization as

$$p(z_1, \ldots, z_N | \boldsymbol{\theta}_m, \mathbf{m}) = \prod_{i=1}^{N} p(z_i | \mathbf{pa}_i, \boldsymbol{\theta}_m, \mathbf{m}) \tag{1}$$

where $\mathbf{Pa}_i$ are the variables corresponding to the parents of $Z_i$ in $\mathbf{m}$. We refer to $p(z_i|\mathbf{pa}_i, \boldsymbol{\theta}_m, \mathbf{m})$ as the *local distribution function* for $Z_i$. Also, when $\mathbf{m}$ appears in an expression $p(\cdot|\cdot)$, it refers to a hypothesis corresponding to the structure $\mathbf{m}$. The hypothesis $\mathbf{m}$ corresponding to Bayesian-network structure $\mathbf{m}$ is the assertion that the structure $\mathbf{m}$ is a perfect map of the joint distribution.

## 2 Embedded Bayesian Network Classifiers

The basic idea behind EBNCs comes from the following observations. Suppose a finite-state node $Y$ has parents $X_1, \ldots, X_n$ as shown in Figure 1a. If each node $X_i$ is binary, there are $2^n$ configurations of $\mathbf{X}$. If $Y$ is also binary, then the traditional local distribution function for $Y$ containing a multinomial distribution for each configuration of $\mathbf{X}$ will contain $2^n$ non-redundant parameters. In contrast, consider the Bayesian network for $Y$ and $\mathbf{X}$ shown in Figure 1b, in which the variables $\mathbf{X}$ are mutually independent given $Y$ (sometimes referred to as a naive-Bayes model). This model contains only $2n+1$ parameters with the traditional encoding of local distribution functions. In addition, as we shall see in Section 4, the local distribution function $p(y|\mathbf{x}, \boldsymbol{\theta}_m, \mathbf{m})$ has a simple form. Therefore, we can imagine using this conditional distribution as a low-dimensional encoding of the conditional distribution for the model in Figure 1a. In general, we have the following definition.

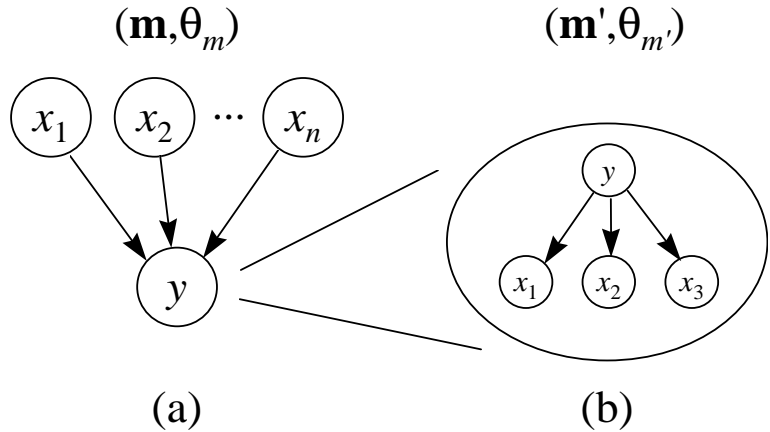

Figure 1: An embedded Bayesian network classifier. The conditional distribution $p(y|\mathbf{x}, \boldsymbol{\theta}_{m'}, \mathbf{m}')$ from the Bayesian network $(\mathbf{m}', \boldsymbol{\theta}_{m'})$ (b) is embedded in the local distribution function $p(y|\mathbf{x}, \boldsymbol{\theta}_y, \mathbf{m})$ for node $Y$ in the Bayesian network $(\mathbf{m}, \boldsymbol{\theta}_m)$ (a).

**Definition** *Let $Y$ be a node in some Bayesian network $(\mathbf{m}, \boldsymbol{\theta}_m)$, and $\mathbf{X}$ be the parents of $Y$ in $(\mathbf{m}, \boldsymbol{\theta}_m)$. Let $(\boldsymbol{\theta}_{m'}, \mathbf{m}')$ be another Bayesian network for precisely the variables $Y$ and $\mathbf{X}$. Then, the local distribution function for $Y$, denoted $p(y|\mathbf{x}, \boldsymbol{\theta}_y, \mathbf{m})$, is said to be an embedded Bayesian network classifier for $Y$ given $\mathbf{X}$ obtained from $(\mathbf{m}, \boldsymbol{\theta}_m)$, if*

$$p(y|\mathbf{x}, \boldsymbol{\theta}_y, \mathbf{m}) = p(y|\mathbf{x}, \boldsymbol{\theta}_{m'}, \mathbf{m}')$$

*and the parameters $\boldsymbol{\theta}_y$ and the parameters associated with other local distribution functions in $(\mathbf{m}, \boldsymbol{\theta}_m)$ are variationally independent.*[1] *The local distribution function is sometimes denoted $EBNC(\mathbf{m}, \boldsymbol{\theta}_y)$ or $EBNC(\mathbf{m}', \boldsymbol{\theta}_{m'})$.*

By this definition, the traditional local distribution function for a finite-state variable having finite-state parents (where each $p(y|\mathbf{x}^j, \boldsymbol{\theta}_y, \mathbf{m})$ is a distinct multinomial distribution) is a special instance of an EBNC. We sometimes refer to the traditional local distribution function as a *trivial* EBNC.

---

[1] Variable $A$ is said to be *variationally dependent* on variable $B$ if knowing the state of $A$ restricts the possible states of $B$.

Although an EBNC can offer a low-dimension encoding of a local distribution function, there are theoretical reasons that make this encoding suspect. Consider again our example where an EBNC is obtained from a naive-Bayes example. In this case, the variables $\mathbf{X}$ are dependent. This dependency is not encoded in the Bayesian network containing the EBNC. In addition, in the naive-Bayes model, the parameters associated with $p(\mathbf{x})$ and $p(y|\mathbf{x})$ are likely to be variationally dependent (Heckerman and Meek, 1997). This dependency is thrown away in the construction of the EBNC. Nonetheless, as we mentioned in the introduction, the use of EBNCs may yield good predictions in practice. In the remainder of this paper, we examine some of the technical details needed to use and learn models containing EBNCs.

## 3 Inference in an EBNC

An EBNC defines a local distribution function $p(y|\mathbf{x}, \boldsymbol{\theta}_y, \mathbf{m})$, which can be computed by performing probabilistic inference in the Bayesian network $(\mathbf{m}', \boldsymbol{\theta}_{m'})$. In this section, we provide a simple formula for computing this function for a given EBNC, assuming all variables in $\mathbf{X}$ are observed. As we shall see, this formula will be useful for demonstrating other useful properties of an EBNC.

Our inference method works in log-odds space. That is, rather than compute $p(y^k|\mathbf{x}, \boldsymbol{\theta}_y, \mathbf{m})$ for all states $y^k$ of $Y$ and all configurations $\mathbf{x}$, we compute

$$\lambda_{k\mathbf{x}} \equiv \log \frac{p(y^k|\mathbf{x}, \boldsymbol{\theta}_y, \mathbf{m})}{p(y^1|\mathbf{x}, \boldsymbol{\theta}_y, \mathbf{m})} \tag{2}$$

for $k = 2, \ldots, r_y$ and all $\mathbf{x}$. This quantity is known as the *posterior log odds of $y^k$ in favor of $y^1$ given $\mathbf{x}$*. Once we compute the $\lambda$s, we transform back to probability space using the *softmax* function:

$$p(y^k|\mathbf{x}, \boldsymbol{\theta}_y, \mathbf{m}) = \frac{e^{\lambda_{k\mathbf{x}}}}{1 + \sum_{j=2}^{r_y} e^{\lambda_{j\mathbf{x}}}} \tag{3}$$

Let $X_1, \ldots, X_{n_h}, Y, X_{n_h+1}, \ldots, X_n$ be a total ordering on the variables that is consistent with $\mathbf{m}$, such that $Y$ appears as late as possible in the ordering. The latter condition says that the node corresponding to $Y$ is an ancestor of each of the nodes corresponding to $X_{n_h+1}, \ldots, X_n$. Given this ordering, we can factor the joint distribution for $Y, X_1, \ldots, X_n$ as follows:

$$p(y, \mathbf{x}|\boldsymbol{\theta}_y, \mathbf{m}) = \left( \prod_{i=1}^{n_h} p(x_i|\mathbf{pa}_i, \boldsymbol{\theta}_y, \mathbf{m}) \right) p(y|\mathbf{pa}_y, \boldsymbol{\theta}_y, \mathbf{m}) \left( \prod_{i=n_h+1}^{n} p(x_i|\mathbf{pa}_i, \boldsymbol{\theta}_y, \mathbf{m}) \right)$$

where $Y$ does not appear in any parent set $\mathbf{Pa}_i$ in the first product. Normalizing to obtain $p(y|\mathbf{x}, \boldsymbol{\theta}_y, \mathbf{m})$, taking a ratio, and canceling like terms, we obtain

$$\lambda_{k\mathbf{x}} = \log \frac{p(y^k|\mathbf{x}, \boldsymbol{\theta}_y, \mathbf{m})}{p(y^1|\mathbf{x}, \boldsymbol{\theta}_y, \mathbf{m})} = \log \frac{\theta(y^k|\mathbf{pa}_y)}{\theta(y^1|\mathbf{pa}_y)} + \sum_{i=n_h+1}^{n} \log \frac{\theta(x_i|\mathbf{pa}_i^k)}{\theta(x_i|\mathbf{pa}_i^1)} \qquad (4)$$

where $\mathbf{pa}_i^k$ is a configuration of $\mathbf{Pa}_i$ in which $y = y^k$, $k = 1, \ldots, r_y$. (Depending on $\mathbf{m}$, some of the terms in the sum may cancel as well.) Equation 4 says that we can determine the posterior log odds simply by summing terms that depend on the configuration of $Y$ and $\mathbf{X}$.

We note that the $\lambda$s can be thought of as parameters for the local distribution function defined by the EBNC. That is, let

$$\boldsymbol{\lambda}_\mathbf{x} \equiv (\lambda_{2\mathbf{x}}, \ldots, \lambda_{r_y\mathbf{x}}) \qquad \mathbf{x} = \mathbf{x}^1, \ldots, \mathbf{x}^{q_y}$$

$$\boldsymbol{\lambda}_y \equiv (\boldsymbol{\lambda}_{\mathbf{x}^1}, \ldots, \boldsymbol{\lambda}_{\mathbf{x}^{q_y}})$$

Then, we can write

$$p(y|\mathbf{x}, \boldsymbol{\theta}_y, \mathbf{m}) = p(y|\mathbf{x}, \boldsymbol{\lambda}_y, \mathbf{m})$$

It is important to note, however, that $\boldsymbol{\lambda}_y$ is not necessarily a set of *free* parameters, because these parameters are derived from the Bayesian network $(\mathbf{m}', \boldsymbol{\theta}_{m'})$. That is, a parameter in $\boldsymbol{\lambda}_y$ may be a (deterministic) function of other parameters in $\boldsymbol{\lambda}_y$.

## 4 The Dimension of an EBNC

In Section 2, we noted that a non-trivial EBNC can be encoded with fewer parameters than the traditional local distribution function, because an EBNC can be encoded with the parameters $\boldsymbol{\theta}_{m'}$. In fact, an EBNC often can be encoded with even fewer parameters. To illustrate this fact, let us consider an EBNC obtained from a the naive Bayesian classifier for $Y$ and $\mathbf{X} = \{X_1, \ldots, X_n\}$ where all variables are binary. (We use $y^1$ and $y^2$ to denote the states of $Y$, and $x_i^1$ and $x_i^2$ to denote the states of each $X_i$.) Using Equation 4, we obtain

$$\log \frac{p(y^2|\mathbf{x}, \boldsymbol{\theta}_y, \mathbf{m})}{p(y^1|\mathbf{x}, \boldsymbol{\theta}_y, \mathbf{m})} = \log \frac{\theta(y^2)}{\theta(y^1)} + \sum_{i=1}^{n} \log \frac{\theta(x_i|y^2)}{\theta(x_i|y^1)} \qquad (5)$$

After some algebra, Equation 5 becomes

$$\log \frac{p(y^2|\mathbf{x}, \boldsymbol{\theta}_y, \mathbf{m})}{p(y^1|\mathbf{x}, \boldsymbol{\theta}_y, \mathbf{m})} = \underbrace{\left( \log \frac{\theta(y^2)}{\theta(y^1)} + \sum_{i=1}^{n} \log \frac{\theta(x_i^1|y^2)}{\theta(x_i^1|y^1)} \right)}_{\equiv \phi_0} + \sum_{i=1}^{n} I(x_i^2) \underbrace{\left( \frac{\theta(x_i^2|y^2)}{\theta(x_i^2|y^1)} - \frac{\theta(x_i^1|y^2)}{\theta(x_i^1|y^1)} \right)}_{\equiv \phi_i}$$
(6)

where $I(x_i^2)$ is the indicator function that is equal to 1 if and only if $x_i = x_i^2$. Equation 6 demonstrates that we can encode the naive EBNC using the parameters $(\phi_0, \ldots, \phi_n)$, which are less in number than the $2n+1$ parameters in $\boldsymbol{\theta}_y$.

This observations raises the question: What is the minimum number of parameters that can be used to encode an EBNC? In the remainder of this section, we provide a procedure for answering this question.

We address this question using the mathematics of differential geometry. A good introduction can be found in Spivak (1979). One way to view our question is to think of the values of $\boldsymbol{\theta}_y$ and $\boldsymbol{\lambda}_y$ as points in Euclidean spaces $R^t$ and $R^l$, respectively. Thus, Equation 2 defines a mapping (i.e., a function) from $R^t$ to $R^l$, which we write as $\boldsymbol{\theta}_y \to \boldsymbol{\lambda}_y$. The set of points in $\boldsymbol{\lambda}_y$ encoded by the EBNC is the image of this function.

Now consider an arbitrary set of points $M$ in $R^l$. The set $M$ is said to be a *d-dimensional manifold in $R^l$* if every point in $M$ possesses a neighborhood that resembles $R^d$—that is, if there is a smooth one-to-one function that maps this neighborhood to an open set in $R^d$ and a smooth inverse mapping. The axes in $R^d$ are often referred to as the *local coordinates* or *local non-redundant parameters* of $M$ in that neighborhood. Sometimes, a single set of coordinates conveniently describe all of $M$—for example, when the mapping from $R^d$ to $M$ is linear. In this case the coordinates are said to form a *global non-redundant set of parameters*. When the mapping is linear, $M$ is said to be a *linear manifold*. In other situations, no one set of coordinates may conveniently represent the image, in which case the manifold is said to be *curved*.

Returning to our problem, we have the following theorem.

**Theorem 1** *Let $EBNC(\mathbf{m}, \boldsymbol{\theta}_y)$ be an embedded Bayesian network classifier for $Y$ and $\mathbf{X}$ obtained from $(\mathbf{m}', \boldsymbol{\theta}_{m'})$. Let $\boldsymbol{\theta}_y \to \boldsymbol{\lambda}_y$ be the mapping defined by this EBNC (Equation 4). Then, the image of this mapping is a linear manifold.*

**Proof:** We can rewrite Equation 4 as follows:

$$\lambda_{k\mathbf{x}} = \log \frac{p(y^k | \mathbf{x}, \boldsymbol{\theta}_y, \mathbf{m})}{p(y^1 | \mathbf{x}, \boldsymbol{\theta}_y, \mathbf{m})} = \underbrace{\log \frac{\theta(y^k | \mathbf{pa}_y)}{\theta(y^1 | \mathbf{pa}_y)}}_{\kappa(y^k | \mathbf{pa}_y)} + \sum_{i=n_h+1}^{n} \underbrace{\log \frac{\theta(x_i | \mathbf{pa}_i^k)}{\theta(x_i | \mathbf{pa}_i^1)}}_{\kappa(x_i | \mathbf{pa}_i^k)} \qquad (7)$$

where we have introduced another set of parameters $\kappa(\cdot|\cdot)$ that we collectively refer to as $\boldsymbol{\kappa}_y$. Equation 7 decomposes the mapping $\boldsymbol{\theta}_y \to \boldsymbol{\lambda}_y$ into the mappings $\boldsymbol{\theta}_y \to \boldsymbol{\kappa}_y \to \boldsymbol{\lambda}_y$ where the first mapping is smooth and many-to-one and the second mapping is linear and many-to-one. Let $\boldsymbol{\phi}_y = (\phi, \ldots, \phi_{d_y})$ be a basis for the image of $\boldsymbol{\kappa}_y \to \boldsymbol{\lambda}_y$. We can find this basis using (e.g.) Gaussian elimination. Now, the image of the mapping $\boldsymbol{\theta}_y \to \boldsymbol{\kappa}_y$ is an open set in $\boldsymbol{\kappa}_y$ because, by definition of hypothesis $\mathbf{m}$, we exclude values of $\boldsymbol{\theta}_y$ that correspond to independencies not encoded in $\mathbf{m}$. Consequently, for every point in the image of the EBNC, there is a smooth one-to-one mapping between the neighborhood of that point and an open subset of $\boldsymbol{\phi}_y$. □

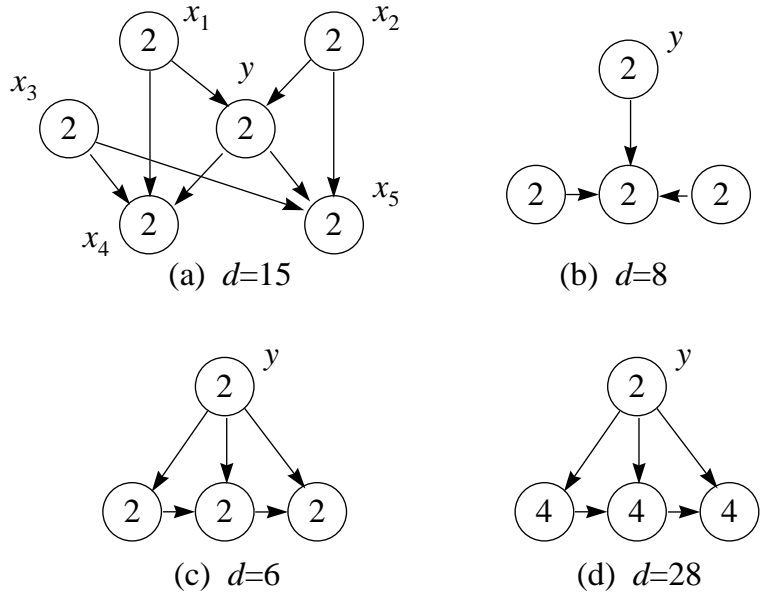

Figure 2: Some Bayesian networks and their EBNC dimensions. The number in the node indicates the number of states of that node.

The proof of Theorem 1 includes a method for finding a set of non-redundant parameters $\boldsymbol{\phi}_y$ for the manifold; and we sometimes denote the model by $\text{EBNC}(\mathbf{m}, \boldsymbol{\phi}_y)$. Figure 2 contains several EBNCs and their corresponding dimensions that we computed using this approach. Unfortunately, the approach is inefficient, because it requires that we apply Gaussian elimination to a matrix that can have up to $2^n$ rows (for $Y$ and $\mathbf{X}$ binary). In Section 6, we consider an alternative construction that is more efficient. First, however, we consider an important ramification of Theorem 1 for learning.

## 5  An Approximation for the Marginal Likelihood

An important quantity for Bayesian learning is the marginal likelihood $p(D|\mathbf{m})$. We do not know of a method for computing the marginal likelihood of a Bayesian network that contains non-trivial EBNCs. Nonetheless, an important consequence of Theorem 1 is that we can apply the work of Haughton (1988) to derive an efficient approximation for the marginal likelihood of a Bayesian network where each local distribution function is an EBNC. Namely, we have the following theorem.

**Theorem 2** *Let $(\mathbf{m}, \boldsymbol{\phi}_m)$ be a Bayesian network for variables $\mathbf{Z} = \{Z_1, \ldots, Z_s\}$ such that the local distribution function for each node $Z_i$ is given by $EBNC(\mathbf{m}_i, \boldsymbol{\theta}_i) = \mathrm{EBNC}(\mathbf{m}_i, \boldsymbol{\phi}_i)$, where $(\mathbf{m}_i, \boldsymbol{\theta}_i)$ is a Bayesian network for $Z_i$ and the parents of $Z_i$ in $\mathbf{m}$. Denote the likelihood of the data associated with the ith variable and its parents by*

$$L_i(\boldsymbol{\phi}_i) = \sum_{l=1}^{N} p(z_{il} | \mathbf{pa}_{il}, \boldsymbol{\phi}_i, \mathbf{m})$$

*where $l$ refers to a case in $D$. Furthermore, assume that the parameter prior $p(\boldsymbol{\phi}_m | \mathbf{m})$ is a probability density function that is non-zero almost everywhere. Then given a complete data set $D$ with $N$ cases,*

$$\log p(D|\mathbf{m}) \approx \log p(\tilde{\boldsymbol{\phi}}_m | \mathbf{m}) + \sum_{i=1}^{s} \left( L_i(\tilde{\boldsymbol{\phi}}_i) + \frac{d_i}{2} \log(2\pi) - \frac{1}{2} \log |A_i| \right) \tag{8}$$

*where $\tilde{\boldsymbol{\phi}}_m$ is the maximum a posteriori (MAP) value for $\boldsymbol{\phi}_m$ given $D$, $d_i$ is the dimension of $EBNC(\mathbf{m}_i, \boldsymbol{\phi}_i)$, and $A_i$ is the negative Hessian of $L_i(\boldsymbol{\phi}_i)$ with respect to $\boldsymbol{\phi}_i$ evaluated at the MAP value for $\boldsymbol{\phi}_i$. The relative error of this approximation is $O_p(N^{-1})$*[2].

**Proof:** Let $\boldsymbol{\lambda}_m = (\boldsymbol{\lambda}_1, \ldots, \boldsymbol{\lambda}_s)$ be the parameters defined by the mappings $\boldsymbol{\phi}_i \to \boldsymbol{\lambda}_i$. The local distribution function for each variable $Z_i$ is a set of multinomial distributions with (not necessarily free) parameters $\boldsymbol{\lambda}_i$. Therefore, if we view the set of variables $\mathbf{Z}$ as a single finite-state variable, then $\mathbf{Z}$ will have a multinomial distribution with (not necessarily free) parameters $\boldsymbol{\gamma}_m(\boldsymbol{\lambda}_m) = (\gamma_2, \ldots, \gamma_t), t = \prod_{i=1}^{s} r_i$, where

$$\log \frac{p(\mathbf{z}^k | \boldsymbol{\gamma}_m, \mathbf{m})}{p(\mathbf{z}^1 | \boldsymbol{\gamma}_m, \mathbf{m})} = \gamma_k$$

and the mapping $\boldsymbol{\lambda}_m \to \boldsymbol{\gamma}_m$ is one-to-one. $\boldsymbol{\gamma}_m$ is sometimes called the *natural parameter space* for the multinomial distribution. Because the parameter sets $\boldsymbol{\phi}_1, \ldots, \boldsymbol{\phi}_s$ are mutually distinct (i.e., variationally independent), it follows from Theorem 1 that the set of values in $\boldsymbol{\lambda}_m$ that are allowed by the local distribution functions $\mathrm{EBNC}(\mathbf{m}_i, \boldsymbol{\phi}_i)$ is a linear manifold of $\boldsymbol{\lambda}_m$ with dimension $d = \sum_{i=1}^{s} d_i$. Furthermore, because the mapping $\boldsymbol{\lambda}_m \to \boldsymbol{\gamma}_m$ is one to one (albeit non-linear), the set of values in $\boldsymbol{\gamma}_m$ that are allowed form a (curved) $d$-dimensional manifold in $\gamma_m$ parameterized globally by $\phi_m$. Consequently, the conditions of Haughton (1988) are satisfied, and we have

$$\log p(D|\mathbf{m}) \approx \log p(\tilde{\phi}_m|\mathbf{m}) + \log p(D|\tilde{\phi}_m, \mathbf{m}) + \left(\sum_{i=1}^{s} d_i \log(2\pi)\right) - \frac{1}{2}\log|A|$$

where $|A|$ is the negative Hessian of the likelihood with respect to $\phi_m$. Finally, because $\phi_1, \ldots, \phi_s$ are mutually distinct, the likelihood $p(D|\phi_m, \mathbf{m})$ factors and the Hessian is block diagonal (one block for each $Z_i$) yielding Equation 8. □

---

[2] Haughton (1988) derives a relative error of $O_p(N^{-1/2})$. The bound we report comes from a personal communication with her.

A more efficient but less accurate approximation is obtained by retaining only those terms in Equation 8 that increase with $N$: $\sum_{i=1}^{s} \log L_i(\tilde{\phi}_i)$, which increases linearly with $N$, and $\sum_{i=1}^{s} \log |A_i|$, which increases as $d_m \log N$. Also, for large $N$, $\tilde{\phi}_m$ can be approximated by the *maximum likelihood* (ML) value of $\phi_m$, $\hat{\phi}_m$, the value of $\phi_m$ for which $p(D|\phi_m, \mathbf{m})$ is a maximum. Thus, we obtain

$$\log p(D|\mathbf{m}) \approx + \sum_{i=1}^{s} \left( L_i(\hat{\phi}_i) - \frac{d_i}{2}\log N \right) \qquad (9)$$

which, under the conditions of Theorem 2, has a relative accuracy of $O_p(1)$.[3] This approximation is called the *Bayesian information criterion* (BIC), and was first derived by Schwarz (1978) for a limited class of models.

The BIC approximation is interesting in several respects. First, it does not depend on the parameter prior, except for the condition that $p(\phi_m|\mathbf{m})$ be non-zero for almost all values of $\phi_m$. Consequently, we can approximate the marginal likelihood without a prior, which is difficult to assess for our models. Second, the approximation is quite intuitive. Namely, it contains a term measuring how well the model with parameters set to an ML value predicts the data $(\log p(D|\hat{\phi}_m, \mathbf{m}))$ and a term that punishes the complexity of the model $(d_m/2 \log N)$. Third, the BIC approximation is exactly the additive inverse of the Minimum Description Length (MDL) criterion described by Rissanen (1987).

When using the Laplace or BIC approximation, we must compute $\tilde{\phi}_m$ or $\hat{\phi}_m$, respectively. Although it may be difficult to determine a global maximum, gradient-based methods such as those described in Gill et al. (1981), Press et al. (1992), and Buntine and Weigand (1994) can be used to locate local maxima. In practice, a good approximation for the marginal likelihood is often obtained by finding many local maxima and then summing their contributions to $p(D|\mathbf{m})$ given by Equation 8 or 9. Note that the ML values $\hat{\phi}_1, \ldots, \hat{\phi}_s$ can be identified separately. Similarly, if the parameter prior factors according to

$$p(\phi_m|\mathbf{m}) = \prod_{i=1}^{s} p(\phi_i|m)$$

---

[3] For some priors $p(\phi_m|\mathbf{m})$, the BIC is accurate to $O(N^{-1/2})$ (Kass and Wasserman, 1995).

then the MAP values $\tilde{\phi}_1, \ldots, \tilde{\phi}_s$ can be identified separately. Buntine and Weigand (1994) and Bishop (1995) survey methods for computing the Hessian terms needed for the Laplace approximation.

An important application of these approximations is the selection of models for classification. In particular, given a class variable $Y$ and input variables $\mathbf{X}$, consider a set of models where the variables in $\mathbf{X}$ are mutually independent, $Y$ is conditioned by each variable in $\mathbf{X}$, and the local distribution functions for $Y$ are obtained from various EBNCs. If we are using the Laplace approximation, assume that $\boldsymbol{\phi}_\mathbf{x}$ and $\boldsymbol{\phi}_y$, the parameters associated with the local distribution functions for $\mathbf{X}$ and $Y$, respectively, are independent, and that the prior distribution for $\boldsymbol{\phi}_\mathbf{x}$ is the same for every model that we consider. If we are using the BIC approximation, assume only that the parameter sets $\boldsymbol{\phi}_\mathbf{x}$ and $\boldsymbol{\phi}_y$ be mutually distinct. Then, the marginal likelihoods for two models will differ only as a result of different local distribution functions for $Y$ given $\mathbf{X}$ and (in the case of the Laplace approximation) the priors $p(\boldsymbol{\phi}_y|\mathbf{m})$.

We note that this procedure provides a method for deciding what subset of the input variables should be used for classification—a task known as *feature selection*. In particular, our procedure may select an EBNC obtained from a Bayesian network that contains a proper subset of the variables $\mathbf{X}$. The corresponding model that we learn for classification will have arcs only from this subset of variables to $Y$. In effect, this subset of inputs (features) has been selected for classification.

## 6 A More Efficient Method for Computing Dimension

In this section, we examine a more efficient method for determining the dimension and a non-redundant parameterization for an EBNC. In so doing, we concentrate on some fixed Bayesian network $(\mathbf{m}, \boldsymbol{\theta}_m)$ from which we obtain the local likelihood. Consequently, we no longer mention $\mathbf{m}$ explicitly in the notation. Furthermore, to avoid difficult notation, we do not give a general construction. Instead, we illustrate our procedure using a specific Bayesian network that is complex enough to capture the essence of the general construction.

Consider the Bayes-net classifier whose structure is shown in Figure 2a. Note that all variables are binary. Given this model structure, the mapping $\boldsymbol{\theta} \to \boldsymbol{\lambda}$ is given by

$$\log \frac{p(y^2|\mathbf{x}, \boldsymbol{\theta})}{p(y^1|\mathbf{x}, \boldsymbol{\theta})} = \log \frac{\theta(y^2|x_1, x_2)}{\theta(y^1|x_1, x_2)} + \log \frac{\theta(x_4|x_1, x_3, y^2)}{\theta(x_4|x_1, x_3, y^1)} + \log \frac{\theta(x_5|x_2, x_3, y^2)}{\theta(x_5|x_2, x_3, y^1)} \quad (10)$$

The overall plan of our construction is to decompose this mapping into a series of mappings $\boldsymbol{\theta} \to \boldsymbol{\eta} \to \boldsymbol{\psi} \to \boldsymbol{\phi} \to \boldsymbol{\lambda}$. In so doing, we will show that $\boldsymbol{\phi}$ is a non-redundant set of parameters for the EBNC. We note that this parameterization may not be the same as the one discussed in Section 4.

Our first step is to decompose $\boldsymbol{\theta} \to \boldsymbol{\lambda}$ into the mappings $\boldsymbol{\theta} \to \boldsymbol{\eta} \to \boldsymbol{\psi} \to \boldsymbol{\lambda}$. We do so by transforming Equation 10 in a manner that generalizes the transformation from Equation 5 to Equation 6. The transformation derives from the observation that if input configurations $\mathbf{x}^1$ and $\mathbf{x}^2$ differ in the state of only one $X_i$, then the difference between the corresponding $\lambda$s will have a simple form. For example, if $\mathbf{x}^1 = (x_1^1, x_2^1, x_3^1, x_4^1, x_5^1)$ and $\mathbf{x}^2$ differs only in $X_1$, then we have

$$\log \frac{p(y^2|x_1^2, x_2^1, x_3^1, x_4^1, x_5^1, \boldsymbol{\theta})}{p(y^1|x_1^2, x_2^1, x_3^1, x_4^1, x_5^1, \boldsymbol{\theta})} - \log \frac{p(y^2|x_1^1, x_2^1, x_3^1, x_4^1, x_5^1, \boldsymbol{\theta})}{p(y^1|x_1^1, x_2^1, x_3^1, x_4^1, x_5^1, \boldsymbol{\theta})} \qquad (11)$$
$$= \left( \log \frac{\theta(y^2|x_1^2, x_2^1)}{\theta(y^1|x_1^2, x_2^1)} - \log \frac{\theta(y^2|x_1^1, x_2^1)}{\theta(y^1|x_1^1, x_2^1)} \right) + \left( \log \frac{\theta(x_4^1|x_1^2, x_3^1, y^2)}{\theta(x_4^1|x_1^2, x_3^1, y^1)} - \log \frac{\theta(x_4^1|x_1^1, x_3^1, y^2)}{\theta(x_4^1|x_1^1, x_3^1, y^1)} \right)$$

Therefore, we compute $\frac{p(y^2|\mathbf{x}, \boldsymbol{\theta})}{p(y^1|\mathbf{x}, \boldsymbol{\theta})}$ for a given $\mathbf{x}$ by first computing $\frac{p(y^2|\mathbf{x}^1, \boldsymbol{\theta})}{p(y^1|\mathbf{x}^1, \boldsymbol{\theta})}$ for the configuration $\mathbf{x}^1$ where each $X_i$ in state $x_i^1$. Then, we sequentially "turn on" each variable $X_i$ that has state $x_i^2$ in $\mathbf{x}$, keeping a running total of difference terms such as the one in Equation 11. For example, if $\mathbf{x} = (x_1^2, x_2^1, x_3^1, x_4^2, x_5^1)$, then we get

$$\log \frac{p(y^2|x_1^2, x_2^1, x_3^1, x_4^2, x_5^1, \boldsymbol{\theta})}{p(y^1|x_1^1, x_2^1, x_3^1, x_4^1, x_5^1, \boldsymbol{\theta})}$$
$$= \log \frac{p(y^2|x_1^1, x_2^1, x_3^1, x_4^1, x_5^1, \boldsymbol{\theta})}{p(y^1|x_1^1, x_2^1, x_3^1, x_4^1, x_5^1, \boldsymbol{\theta})}$$
$$+ \left\{ \log \frac{p(y^2|x_1^2, x_2^1, x_3^1, x_4^1, x_5^1, \boldsymbol{\theta})}{p(y^1|x_1^2, x_2^1, x_3^1, x_4^1, x_5^1, \boldsymbol{\theta})} - \log \frac{p(y^2|x_1^1, x_2^1, x_3^1, x_4^1, x_5^1, \boldsymbol{\theta})}{p(y^1|x_1^1, x_2^1, x_3^1, x_4^1, x_5^1, \boldsymbol{\theta})} \right\}$$
$$+ \left\{ \log \frac{p(y^2|x_1^2, x_2^1, x_3^1, x_4^2, x_5^1, \boldsymbol{\theta})}{p(y^1|x_1^2, x_2^1, x_3^1, x_4^2, x_5^1, \boldsymbol{\theta})} - \log \frac{p(y^2|x_1^2, x_2^1, x_3^1, x_4^1, x_5^1, \boldsymbol{\theta})}{p(y^1|x_1^2, x_2^1, x_3^1, x_4^1, x_5^1, \boldsymbol{\theta})} \right\}$$
$$= \log \frac{p(y^2|x_1^1, x_2^1, x_3^1, x_4^1, x_5^1, \boldsymbol{\theta})}{p(y^1|x_1^1, x_2^1, x_3^1, x_4^1, x_5^1, \boldsymbol{\theta})}$$
$$+ \left\{ \left( \log \frac{\theta(y^2|x_1^2, x_2^1)}{\theta(y^1|x_1^2, x_2^1)} - \log \frac{\theta(y^2|x_1^1, x_2^1)}{\theta(y^1|x_1^1, x_2^1)} \right) + \left( \log \frac{\theta(x_4^1|x_1^2, x_3^1, y^2)}{\theta(x_4^1|x_1^2, x_3^1, y^1)} - \log \frac{\theta(x_4^1|x_1^1, x_3^1, y^2)}{\theta(x_4^1|x_1^1, x_3^1, y^1)} \right) \right\}$$
$$+ \left\{ \log \frac{\theta(x_4^2|x_1^2, x_3^1, y^2)}{\theta(x_4^2|x_1^2, x_3^1, y^1)} - \log \frac{\theta(x_4^1|x_1^2, x_3^1, y^2)}{\theta(x_4^1|x_1^2, x_3^1, y^1)} \right\}$$

For an arbirary input configuration $\mathbf{x}$, we obtain

$$\log \frac{p(y^2|x_1, x_2, x_3, x_4, x_5, \boldsymbol{\theta})}{p(y^1|x_1, x_2, x_3, x_4, x_5, \boldsymbol{\theta})} =$$
$$\underbrace{\eta_0}_{\equiv \psi_0}$$

$$+I(x_1^2) \left\{ \underbrace{\eta_1(y|x_1, x_2^1) + \eta_1(x_4^1|x_1, x_3^1, y)}_{\equiv \psi_1} \right\}$$

$$+I(x_2^2) \left\{ I(x_1^1) \left[ \underbrace{\eta_2(y|x_1^1, x_2) + \eta_2(x_5^1|x_2, x_3^1, y)}_{\equiv \psi_2(x_1^1)} \right] \right.$$

$$\left. + I(x_1^2) \left[ \underbrace{\eta_2(y|x_1^2, x_2) + \eta_2(x_5^1|x_2, x_3^1, y)}_{\equiv \psi_2(x_1^2)} \right] \right\}$$

$$+I(x_3^2) \left\{ I(x_1^1) \ I(x_2^1) \left[ \underbrace{\eta_3(x_4^1|x_1^1, x_3, y) + \eta_3(x_5^1|x_2^1, x_3, y)}_{\equiv \psi_3(x_1^1, x_2^1)} \right] \right.$$

$$+I(x_1^2) \ I(x_2^1) \left[ \underbrace{\eta_3(x_4^1|x_1^2, x_3, y) + \eta_3(x_5^1|x_2^1, x_3, y)}_{\equiv \psi_3(x_1^2, x_2^1)} \right]$$

$$+I(x_1^1) \ I(x_2^2) \left[ \underbrace{\eta_3(x_4^1|x_1^1, x_3, y) + \eta_3(x_5^1|x_2^2, x_3, y)}_{\equiv \psi_3(x_1^1, x_2^2)} \right]$$

$$\left. + I(x_1^2) \ I(x_2^2) \left[ \underbrace{\eta_3(x_4^1|x_1^2, x_3, y) + \eta_3(x_5^1|x_2^2, x_3, y)}_{\equiv \psi_3(x_1^2, x_2^2)} \right] \right\}$$

$$+I(x_4^2) \left\{ I(x_1^1) \ I(x_3^1) \underbrace{\eta_4(x_4|x_1^1, x_3^1, y)}_{\equiv \psi_4(x_1^1, x_3^1)} \right.$$

$$+I(x_1^2) \ I(x_3^1) \underbrace{\eta_4(x_4|x_1^2, x_3^1, y)}_{\equiv \psi_4(x_1^2, x_3^1)}$$

$$+I(x_1^1) \ I(x_3^2) \underbrace{\eta_4(x_4|x_1^1, x_3^2, y)}_{\equiv \psi_4(x_1^1, x_3^2)}$$

$$\left. + I(x_1^2) \ I(x_3^2) \underbrace{\eta_4(x_4|x_1^2, x_3^2, y)}_{\equiv \psi_4(x_1^2, x_3^2)} \right\}$$

$$+I(x_5^2) \left\{ I(x_2^1) \ I(x_3^1) \underbrace{\eta_5(x_5|x_2^1, x_3^1, y)}_{\equiv \psi_5(x_2^1, x_3^1)} \right.$$

$$+ I(x_2^2) \ I(x_3^1) \ \underbrace{\eta_5(x_5|x_2^2,x_3^1,y)}_{\equiv \psi_5(x_2^2,x_3^1)}$$

$$+ I(x_2^1) \ I(x_3^2) \ \underbrace{\eta_5(x_5|x_2^1,x_3^2,y)}_{\equiv \psi_5(x_2^1,x_3^2)}$$

$$+ I(x_2^2) \ I(x_3^2) \ \underbrace{\eta_5(x_5|x_2^2,x_3^2,y)}_{\equiv \psi_5(x_2^2,x_3^2)} \Bigg\} \quad (12)$$

where $I(x_i^j)$ is the indicator function that is equal to 1 if $x_i = x_i^j$ and 0 otherwise. In Equation 12 we have introduced the parameters $\boldsymbol{\eta}$ defined by

$$\eta_0 \equiv \log \frac{p(y^2|x_1^1,x_2^1,x_3^1,x_4^1,x_5^1,\boldsymbol{\theta})}{p(y^1|x_1^1,x_2^1,x_3^1,x_4^1,x_5^1,\boldsymbol{\theta})} \quad (13)$$

$$\eta_i(y|x_i,\mathbf{a}) \equiv \log \frac{\theta(y^2|x_i^2,\mathbf{a})}{\theta(y^1|x_i^2,\mathbf{a})} - \log \frac{\theta(y^2|x_i^1,\mathbf{a})}{\theta(y^1|x_i^1,\mathbf{a})} \quad (14)$$

$$\eta_i(x_i|\mathbf{a},y) \equiv \log \frac{\theta(x_i^2|\mathbf{a},y^2)}{\theta(x_i^2|\mathbf{a},y^1)} - \log \frac{\theta(x_i^1|\mathbf{a},y^2)}{\theta(x_i^1|\mathbf{a},y^1)} \quad (15)$$

$$\eta_i(x_j^k|x_i,\mathbf{a},y) \equiv \log \frac{\theta(x_j^k|x_i^2,\mathbf{a},y^2)}{\theta(x_j^k|x_i^2,\mathbf{a},y^1)} - \log \frac{\theta(x_j^k|x_i^1,\mathbf{a},y^2)}{\theta(x_j^k|x_i^1,\mathbf{a},y^1)} \quad k=1,2 \quad (16)$$

where $\mathbf{a}$ is any (possibly empty) configuration of variables excluding $X_i$ and $Y$. In addition, we have introduced the parameters $\boldsymbol{\psi}$, where $\psi_i(\mathbf{a})$ is the coefficient of the product $I(x_i^2)I(\mathbf{a})$. We use $\eta_i$ and $\psi_i$ to denote the collection of parameters $\eta_i(\cdot|\cdot)$ and $\psi_i(\cdot)$, respectively.

Equation 12 defines a series of mappings $\boldsymbol{\theta} \to \boldsymbol{\eta} \to \boldsymbol{\psi} \to \boldsymbol{\lambda}$ that we can use to determine the dimension of the model as follows. First, note that the mappings $\boldsymbol{\eta} \to \boldsymbol{\psi} \to \boldsymbol{\lambda}$ are linear. Therefore, if the image of the mapping $\boldsymbol{\theta} \to \boldsymbol{\eta}$ is open in $\boldsymbol{\eta}$, then the dimension of the EBNC will be

$$d = \text{rank}\left[\frac{\partial \boldsymbol{\lambda}}{\partial \boldsymbol{\eta}}\right] \quad (17)$$

where $\frac{\partial \boldsymbol{\lambda}}{\partial \boldsymbol{\eta}}$ is the Jacobian matrix of the linear mapping $\boldsymbol{\eta} \to \boldsymbol{\lambda}$.

Given the form of the mappings $\boldsymbol{\eta} \to \boldsymbol{\psi} \to \boldsymbol{\lambda}$, we can perform row reductions on the Jacobian in Equation 17 to obtain a block diagonal matrix, where block $i$ corresponds to $\frac{\partial \psi_i}{\partial \eta_i}$, the Jacobian matrix for the mapping $\eta_i$ to parameters $\psi_i$. Consequently, we have

$$d = \sum_{i=1}^{n} \text{rank}\left[\frac{\partial \psi_i}{\partial \eta_i}\right] \quad (18)$$

For the model in Figure 2d, we obtain $d = 1+1+2+3+4+4 = 15$, which we also obtain using the method described in Section 4.

Using the form of the mappings $\boldsymbol{\eta} \to \boldsymbol{\psi} \to \boldsymbol{\lambda}$, we also can obtain a non-redundant set of parameters for the EBNC in a straightforward manner. Namely, let basis$(\eta_i \to \psi_i)$ be a basis for the image of the mapping $\eta_i \to \psi_i$. Then, because the parameter sets $\psi_1, \ldots, \psi_n$ do not overlap,

$$\boldsymbol{\phi} = \cup_{i=1}^n \text{ basis}(\eta_i \to \psi_i) \qquad (19)$$

is a set of non-redundant parameters for the EBNC.

Finally, we need to show that the image of the mapping $\boldsymbol{\theta} \to \boldsymbol{\eta}$ is open in $\boldsymbol{\eta}$. To do so, we decompose this mapping into two mappings $\boldsymbol{\theta} \to \boldsymbol{\omega} \to \boldsymbol{\eta}$, where

$$\omega_0 \equiv \eta_0 \qquad (20)$$

$$\omega(y|x_i^k, \mathbf{a}) \equiv \log \frac{\theta(y^2|x_i^k, \mathbf{a})}{\theta(y^1|x_i^k, \mathbf{a})} \qquad (21)$$

$$\omega(x_i^k|\mathbf{a}, y) \equiv \log \frac{\theta(x_i^k|\mathbf{a}, y^2)}{\theta(x_i^k|\mathbf{a}, y^1)} \qquad (22)$$

$$\omega(x_j^l|x_i^k, \mathbf{a}, y) \equiv \log \frac{\theta(x_j^l|x_i^k, \mathbf{a}, y^2)}{\theta(x_j^l|x_i^k, \mathbf{a}, y^1)} \qquad (23)$$

for $k = 1, 2$ and $l = 1, 2$. The image of the mapping $\boldsymbol{\theta} \to \boldsymbol{\omega}$ is open in $\boldsymbol{\omega}$, because any two parameters in $\boldsymbol{\omega}$ are either functions of different parameters in $\boldsymbol{\theta}$, or are of the form

$$\omega(w_i^1|\mathbf{a}, y) = \log \frac{\theta(x_i^1|\mathbf{a}, y^2)}{\theta(x_i^1|\mathbf{a}, y^1)}$$

$$\omega(w_i^2|\mathbf{a}, y) = \log \frac{1 - \theta(x_i^1|\mathbf{a}, y^2)}{1 - \theta(x_i^1|\mathbf{a}, y^1)}$$

where $\omega(w_i^1|\mathbf{a}, y)$ can be varied independently of $\omega(w_i^2|\mathbf{a}, y)$ and vice versa. Furthermore, the Jacobian matrix of the linear mapping $\boldsymbol{\omega} \to \boldsymbol{\eta}$ can be made triangular. Consequently, the imagine of the combined mapping $\boldsymbol{\theta} \to \boldsymbol{\eta}$ is open in $\boldsymbol{\eta}$.

The generalization of Equations 18 and 19 to an arbitrary Bayes-net classifier with binary variables is straightforward. First, we write down the mapping $\boldsymbol{\theta} \to \boldsymbol{\lambda}$ analogous to Equation 10. We then use the factorization in this mapping to sequentially decompose $\frac{p(y^2|\mathbf{x}, \boldsymbol{\theta})}{p(y^1|\mathbf{x}, \boldsymbol{\theta})}$ as in Equation 12. (We can use any variable order to build the decomposition, although we have found that the computations are most efficient when we use the ordering consistent with the Bayesian-network structure.) This decomposition yields mappings analogous to $\boldsymbol{\eta} \to \boldsymbol{\psi} \to \boldsymbol{\lambda}$ in our example. Finally, we compute a basis for the image of each mapping $\eta_i \to \psi_i$ using (e.g.) Gaussian elimination. The generalization to non-binary variables involves additional book keeping, but the form of the basic construction remains the same.

The worst-case computational complexity of this procedure is exponential in $n$. In particular, some sets $\psi_i$ may contain $O(2^n)$ parameters (assuming variables are binary). Consequently, a computation of the basis will have computational complexity $O(2^{3n})$. Nonetheless our construction is tractable in practice.

We can use our procedure to determine the dimension and a non-redundant set of parameters for an EBNC having various canonical forms. For example, consider the EBNC obtained from a Bayesian classifier where $Y$ is a root node and the inputs $X_1, \ldots, X_n$ form a Markov chain conditioned on $Y$. Using our method, we obtain

$$\log \frac{p(y^2|\mathbf{x},\boldsymbol{\theta})}{p(y^1|\mathbf{x},\boldsymbol{\theta})}$$

$$= \underbrace{\eta_0}_{\equiv \eta_i} + I(x_i^2) \underbrace{\eta_1(x_1|y)}_{\equiv \psi_1} + \sum_{i=2}^n I(x_i^2) \left\{ I(x_{i-1}^1) \underbrace{\eta_i(x_i|x_{i-1}^1, y)}_{\equiv \psi_i(x_{i-1}^1)} + I(x_{i-1}^2) \underbrace{\eta_i(x_i|x_{i-1}^2, y)}_{\equiv \psi_i(x_{i-1}^2)} \right\}$$

Consequently, $d = 1 + 1 + 2(n-1) = 2n$.

Also, we can use the mapping $\boldsymbol{\phi}_y \to \boldsymbol{\lambda}_y$ as an alternative inference method to determine the local distribution function $p(y|\mathbf{x}, \boldsymbol{\phi}_y, \mathbf{m})$. In addition, we note that Equation 3 and the generalization of Equation 12 demonstrate that an EBNC is a special case of a softmax polynomial regression. Finally, we note that our work can be generalized to include situations where some of the domain variables are continuous.

## Acknowledgments

We thank Steve Altschuler, Max Chickering, Robert Gutschera, and Lani Wu for useful discussions.